# Densely Connected Bidirectional LSTM with Applications to Sentence Classification

Zixiang Ding[1], Rui Xia[1] *, Jianfei Yu[2], Xiang Li[1], Jian Yang[1]

[1] School of Computer Science and Engineering, Nanjing University of Science and Technology
[2] School of Information Systems, Singapore Management University
{dingzixiang, rxia}@njust.edu.cn, jfyu.2014@phdis.smu.edu.sg,
{xiang.li.implus, csjyang}@njust.edu.cn

## Abstract

Deep neural networks have recently been shown to achieve highly competitive performance in many computer vision tasks due to their abilities of exploring in a much larger hypothesis space. However, since most deep architectures like stacked RNNs tend to suffer from the vanishing-gradient and overfitting problems, their effects are still understudied in many NLP tasks. Inspired by this, we propose a novel multi-layer RNN model called densely connected bidirectional long short-term memory (DC-Bi-LSTM) in this paper, which essentially represents each layer by the concatenation of its hidden state and all preceding layers' hidden states, followed by recursively passing each layer's representation to all subsequent layers. We evaluate our proposed model on five benchmark datasets of sentence classification. DC-Bi-LSTM with depth up to 20 can be successfully trained and obtain significant improvements over the traditional Bi-LSTM with the same or even less parameters. Moreover, our model has promising performance compared with the state-of-the-art approaches.

## 1 Introduction

With the recent trend of deep learning, various kinds of deep neural architectures have been proposed for many tasks in speech recognition [Graves *et al.*, 2013], computer vision [Russakovsky *et al.*, 2015] and natural language processing (NLP) [Irsoy and Cardie, 2014], which have been shown to achieve better performance than both traditional methods and shallow architectures. However, since conventional deep architectures often suffer from the well-known vanishing-gradient and overfitting problems, most of them are not easy to train and therefore cannot achieve very satisfactory performance.

To address these problems, different approaches have been recently proposed for various computer vision tasks, including Highway Networks [Srivastava *et al.*, 2015], ResNet [He *et al.*, 2016] and GoogLeNet [Szegedy *et al.*, 2015;

---
*The corresponding author of this paper.

2016]. One of the representative work among them is the recently proposed Dense Convolutional Networks (DenseNet) [Huang *et al.*, 2017]. Different from previous work, to strengthen information flow between layers and reduce the number of parameters, DenseNet proposes to directly connect all layers in a feed-forward fashion and encourages feature reuse through representing each layer by concatenating the feature-maps of all preceding layers as input. Owing to this well-designed densely connected architecture, DenseNet obtains significant improvements over the state-of-the-art results on four highly competitive object recognition benchmark tasks (CIFAR-10, CIFAR-100, SVHN, and ImageNet).

Motivated by these successes in computer vision, some deep architectures have also been recently applied in many NLP applications. Since recurrent neural networks (RNNs) are effective to capture the flexible context information contained in texts, most of these deep models are based on the variants of RNNs. Specifically, on basis of Highway Networks, Zhang *et al.* [2016b] proposed Highway LSTM to extend stacked LSTM by introducing gated direct connections between memory cells in adjacent layers. Inspired by ResNet, Yu *et al.* [2017] further proposed a hierarchical LSTM enhanced by residual learning for relation detection task. However, to the best of our knowledge, the application of DenseNet to RNN has not been explored in any NLP task before, which is the motivation of our work.

Therefore, in this paper, we propose a novel multi-layer RNN model called Densely Connected Bidirectional LSTM (DC-Bi-LSTM) for sentence classification. The architecture of DC-Bi-LSTM is shown in Figure 1, where we use Bi-LSTM to encode the input sequence, and regard the sequence of hidden states as reading memory for each layer. In detail, we obtain first-layer reading memory based on original input sequence, and second-layer reading memory based on the position-aligned concatenation of original input sequence and first-layer reading memory, and so on. Finally, based on the concatenation of original input sequence and all previous reading memory, we get the $n$-th-layer reading memory, which is then taken as the final feature representation for classification.

We evaluate our proposed architecture on five sentence classification datasets, including Movie Review Data [Pang and Lee, 2005] and Stanford Sentiment Tree-bank [Socher *et al.*, 2013] for fine-grained and polarity sentiment classifica-tions, TREC dataset [Li and Roth, 2002] for question type classification and subjectivity classification dataset [Pang and Lee, 2004]. DC-Bi-LSTM with depth up to 20 can be successfully trained and significantly outperform the traditional Bi-LSTM with the same or even less parameters. Moreover, our model achieves indistinguishable performance in comparison with the state-of-the-art approaches.

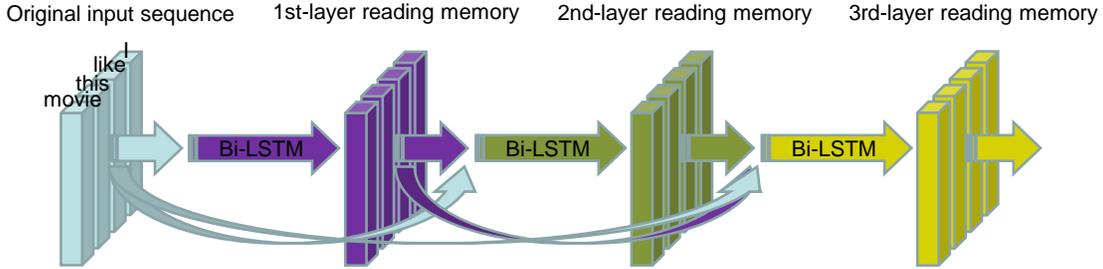

Figure 1: The architecture of DC-Bi-LSTM. We obtain first-layer reading memory based on original input sequence, and second-layer reading memory based on the position-aligned concatenation of original input sequence and first-layer reading memory, and so on. Finally, we get the $n$-th-layer reading memory and take it as the final feature representation for classification.

The main contributions of our work are summarized as follows:

- We propose a novel deep RNNs architecture called DC-Bi-LSTM. Compared with conventional deep stacked RNNs, DC-Bi-LSTM alleviates the problems of vanishing-gradient and overfitting and can be successfully trained when the networks are as deep as dozens of layers.
- We conducted experiments on five datasets of sentence classification, our model obtains significant improvements over the traditional Bi-LSTM and gets promising performance in comparison with the state-of-the-art approaches.

## 2 Related Work

### 2.1 Sentence Classification

The challenge for sentence classification is to perform compositions over variable-length sentences and capture useful features for classification.

Traditional methods are commonly based on the bag-of-words (BoW) model, which treats sentences as unordered collections thus fails to capture syntactic structures and contextual information. Recently, deep learning has developed rapidly in natural language processing, the first breakthrough is learning word embedding [Bengio *et al.*, 2003; Mikolov *et al.*, 2013]. With the help of word embedding, some composition based methods are proposed. For example, Recursive Neural Networks [Socher *et al.*, 2013; Irsoy and Cardie, 2014] build representations of phrases and sentences by combining neighboring constituents based on the parse tree. Convolutional Neural Networks [Kim, 2014] use convolutional filters to extract local features over word embedding matrices. RNNs with Long Short-Term Memory units [Mikolov, 2012; Chung *et al.*, 2014; Tai *et al.*, 2015]are effective networks to process sequential data, which analyze a text word by word and stores the semantics of all the previous text in a fixed-sized hidden state. In this way, LSTM can better capture the contextual information and semantics of long texts. Moreover, bidirectional RNNs [Schuster and Paliwal, 1997] processes the sequence both forward and backward, naturally, a better semantic representation can usually be obtained than unidirectional RNNs.

### 2.2 Stacked RNNs and Extensions

Schmidhuber [1992]; EI Hihi and Bengio [1996] introduced stacked RNNs by stacking RNN layers on top of each other. The hidden states of RNN below are taken as inputs to the RNN above. However, it is very hard to train deep stacked RNNs due to the feed-forward structure of stacked layers. Below are some extensions to alleviate this problem.

Skip connections (or shortcut connections) enable unimpeded information flow by adding direct connections across different layers thus alleviate the gradient problems. For example, Raiko et al. (2012); Graves (2013); Hermans and Schrauwen (2013); Wu et al. (2016); Yu et al. (2017) introduced skip connections into stacked RNNs and make it easier to build deeper stacked RNNs. In addition to using skip connections, Yao et al. (2015) and Zhang et al. (2016) proposed highway LSTMs, which extend stacked RNNs by introducing gated direct connections between memory cells in adjacent layers.

## 3 Model

In this section, we describe the architecture of our proposed Densely Connected Bidirectional LSTM (DC-Bi-LSTM) model for sentence classification.

### 3.1 Long Short-Term Memory

Given an arbitrary-length input sentence $S = \{w_1, w_2, \ldots, w_s\}$, Long Short-Term Memory (LSTM) [Hochreiter and Schmidhuber, 1997] computes the hidden states $h = \{h_1, h_2, \ldots, h_s\}$ by iterating the following equations:

$$h_t = \text{lstm}(h_{t-1}, e(w_t)). \quad (1)$$

The detailed computation is described as follows:

$$\begin{pmatrix} i \\ f \\ o \\ g \end{pmatrix} = \begin{pmatrix} \text{sigm} \\ \text{sigm} \\ \text{sigm} \\ \tanh \end{pmatrix} T_{m+d, 4d}([e(w_t); h_{t-1}]), \quad (2)$$

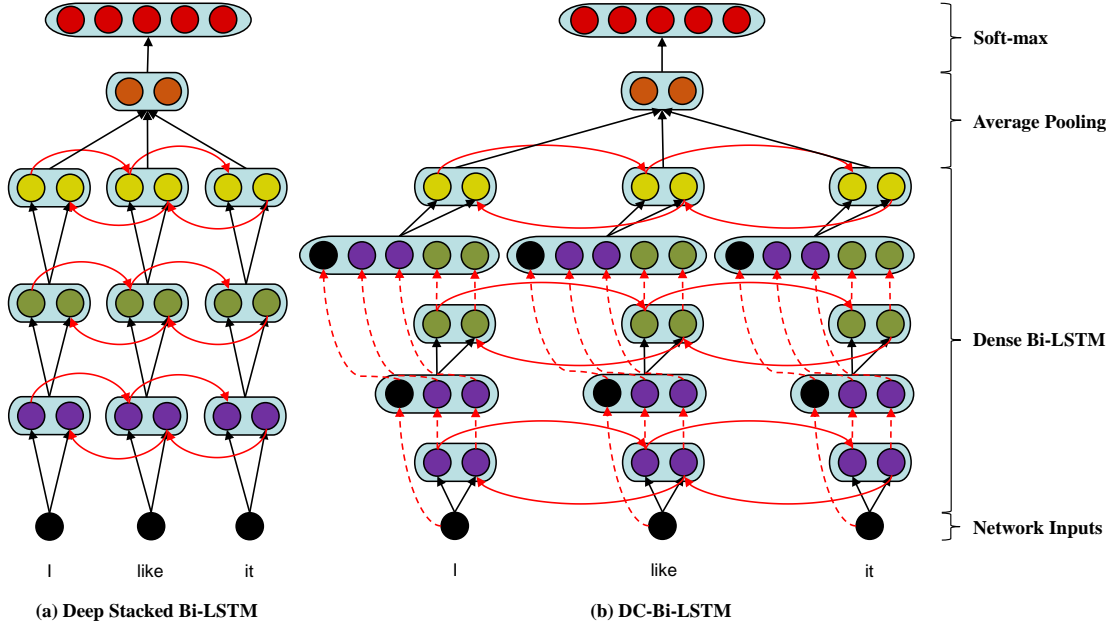

Figure 2: Illustration of (a) Deep Stacked Bi-LSTM and (b) DC-Bi-LSTM. Each black node denotes an input layer. Purple, green, and yellow nodes denote hidden layers. Orange nodes denote average pooling of forward or backward hidden layers. Each red node denotes a class. Ellipse represents the concatenation of its internal nodes. Solid lines denote the connections of two layers. Finally, dotted lines indicate the operation of copying.

$$c_t = f \odot c_{t-1} + i \odot g, \quad (3)$$
$$h_t = o \odot \tanh(c_t), \quad (4)$$

where $e(w_t) \in R^m$ is the word embedding of $w_t$, $h_{t-1} \in R^d$ is the hidden state of LSTM at time step $t-1$, $[e(w_t); h_{t-1}] \in R^{m+d}$ is concatenation of the two vectors. $T_{m+d,4d} : R^{m+d} \to R^{4d}$ is an affine transform ($Wx + b$ for some $W$ and $b$). sigm and tanh are respectively sigmoid and hyperbolic tangent activation functions, $i, f, o, g \in R^d$ are respectively input gate, forget gate, output gate and new candidate memory state. Particularly, $c_t \in R^d$ is additional memory cell used for capturing long distance dependencies, $\odot$ denotes element-wise multiplication.

### 3.2 Deep Stacked Bi-LSTM

Bidirectional LSTM (Bi-LSTM) [Graves *et al.*, 2013] uses two LSTMs to process sequence in two directions: forward and backward. In this way, the forward and backward contexts can be considered simultaneously. The calculations of Bi-LSTMs can be formulated as follows:

$$\overrightarrow{h_t} = \text{lstm}(\overrightarrow{h_{t-1}}, e(w_t)), \quad (5)$$
$$\overleftarrow{h_t} = \text{lstm}(\overleftarrow{h_{t-1}}, e(w_t)). \quad (6)$$

Then the concatenation of forward and backward hidden states is taken as the representation of each word. For word $w_t$, the representation is denoted as $h_t = [\overrightarrow{h_t}; \overleftarrow{h_t}]$.

As shown in Figure 2 (a), deep stacked Bi-LSTM [Schmidhuber, 1992]; [El Hihi and Bengio, 1996] uses multiple Bi-LSTMs with different parameters in a stacking way. The hidden state of $l$-layer Bi-LSTM can be represented as $h_t^l$, which is the concatenation of forward hidden state $\overrightarrow{h_t^l}$ and backward hidden state $\overleftarrow{h_t^l}$. The calculation of $h_t^l$ is as follows:

$$h_t^l = [\overrightarrow{h_t^l}; \overleftarrow{h_t^l}], \text{ specially, } h_t^0 = e(w_t), \quad (7)$$
$$\overrightarrow{h_t^l} = \text{lstm}(\overrightarrow{h_{t-1}^l}, h_t^{l-1}), \quad (8)$$
$$\overleftarrow{h_t^l} = \text{lstm}(\overleftarrow{h_{t+1}^l}, h_t^{l-1}). \quad (9)$$

### 3.3 Densely Connected Bi-LSTM

As shown in Figure 2 (b), Densely Connected Bi-LSTM (DC-Bi-LSTM) consists of four modules: network inputs, dense Bi-LSTM, average pooling and soft-max layer.

**(1) Network Inputs**

The input of our model is a variable-length sentence, which can be represented as $S = \{w_1, w_2, \ldots, w_s\}$. Like other deep learning models, each word is represented as a dense vector extracted from a word embedding matrix. Finally, a sequence of word vectors $\{e(w_1), e(w_2), \ldots, e(w_s)\}$ is sent to the dense Bi-LSTM module as inputs.

**(2) Dense Bi-LSTM**

This module consists of multiple Bi-LSTM layers. For the first Bi-LSTM layer, the input is a word vector sequence $\{e(w_1), e(w_2), \ldots, e(w_s)\}$, and the output is $h^1 = \{h_1^1, h_2^1, \ldots, h_s^1\}$, in which $h_t^1 = [\overrightarrow{h_t^1}; \overleftarrow{h_t^1}]$ as described in Section 3.2. For the second Bi-LSTM layer, the input is not the sequence $\{h_1^1, h_2^1, \ldots, h_s^1\}$ (the way stacked RNNs use), but the concatenation of all previous outputs, formulated as $\{[e(w_1); h_1^1], [e(w_2); h_2^1], \ldots, [e(w_s); h_s^1]\}$, and the output is $h^2 = \{h_1^2, h_2^2, \ldots, h_s^2\}$. For the third layer, whose input is $\{[e(w_1); h_1^1; h_1^2], [e(w_2); h_2^1; h_2^2], \ldots, [e(w_s); h_s^1; h_s^2]\}$, like the second layer does. The rest layers process similarly and omitted for brevity. The above process is formulated as follows:

$$h_t^l = [\overrightarrow{h_t^l}; \overleftarrow{h_t^l}], \text{ specially, } h_t^0 = e(w_t), \quad (10)$$

$$\overrightarrow{h_t^l} = \text{lstm}(\overrightarrow{h_{t-1}^l}, M_t^{l-1}), \quad (11)$$

$$\overleftarrow{h_t^l} = \text{lstm}(\overleftarrow{h_{t+1}^l}, M_t^{l-1}), \quad (12)$$

$$M_t^{l-1} = [h_t^0; h_t^1; \ldots; h_t^{l-1}]. \quad (13)$$

**(3) Average Pooling**

For a $L$ layer Dense Bi-LSTM, the output is $h^L = \{h_1^L, h_2^L, \ldots, h_s^L\}$. Average pooling module reads in $h^L$ and calculate the average value of these vectors, the computation can be formulated as $h^* = \text{average}(h_1^L, h_2^L, \ldots, h_s^L)$.

**(4) Soft-max Layer**

This module is a simple soft-max classifier, which takes $h^*$ as features and generates predicted probability distribution over all sentence labels.

### 3.4 Comparison with Deep Stacked Bi-LSTM

As shown in Figure 2, they have the same network inputs, average pooling and soft-max layer, but differ in dense Bi-LSTM module. For the $k$-th Bi-LSTM layer, the input of deep stacked Bi-LSTM is $\{h_1^{k-1}, h_2^{k-1}, \ldots, h_s^{k-1}\}$, while for densely connected Bi-LSTM, the input is $\{[e(w_1); h_1^1; h_1^2; \ldots; h_1^{k-1}], [e(w_2); h_2^1; h_2^2; \ldots; h_2^{k-1}], \ldots, [e(w_s); h_s^1; h_s^2; \ldots; h_s^{k-1}]\}$. Thanks to this densely connected structure, DC-Bi-LSTM have several advantages:

- Easy to train even when the network is very deep. The reason is that: for every RNN layer, the output is directly sent to the last RNN layer as input and leads to an implicit deep supervision, which alleviates the problem of vanishing-gradient.

- Better parameter efficiency, that means, DC-Bi-LSTM obtains better performance with equal or less parameters compared with traditional RNNs or deep stacked RNNs. That is because: for every RNN layer, it can directly read the original input sequence thus it doesn't have the burden to pass on all useful information and just adds information to the network. Therefore, DC-Bi-LSTM layers are very narrow (for example, 10 hidden units per layer).

It is worth to note that deep residual Bi-LSTM [Yu *et al.*, 2017] looks very similar to our model, however, they are essentially different. For each layer, deep residual Bi-LSTM uses point-wise summation to merge input into output, which may impede the information flow in the network. In contrast, our model merges input into output by concatenation to further improve the information flow.

Table 2: Summary statistics of benchmark datasets. c: Number of target classes. l: Average sentence length. train/dev/test: size of train/development/test set, CV in test column means 10-fold cross validation.

| Data | c | l | train | dev | test |
|------|---|----|-------|------|------|
| MR   | 2 | 20 | 10662 | -    | CV   |
| SST-1| 5 | 18 | 8544  | 1101 | 2210 |
| SST-2| 2 | 19 | 6920  | 872  | 1821 |
| Subj | 2 | 23 | 10000 | -    | CV   |
| TREC | 6 | 10 | 5452  | -    | 500  |

### 3.5 Potential Application Scenario

From a semantic perspective, the dense Bi-LSTM module adds multi-read context information of each word into their original word vector in a concatenation way: $h^1$ is the first reading memory based on the input sentence $S$, $h^2$ is the second reading memory based on $S$ and $h^1$, $h^k$ is the $k$-th reading memory based on $S$ and all previous reading memory. Since the word vector for each word is completely preserved, this module is harmless and can be easily added to other models that use RNN. For example, in the task of machine translation and dialog system, the Bi-LSTM encoder can be replaced by dense Bi-LSTM module and may bring improvements.

## 4 Experiments
### 4.1 Dataset

DC-Bi-LSTM are evaluated on several benchmark datasets, and the summary statistics are shown in Table 2.

- MR: Movie Review Data is a popular sentiment classification dataset proposed by [Pang and Lee, 2005]. Each review belongs to positive or negative sentiment class and contains only one sentence.

- SST-1: Stanford Sentiment Treebank is an extension of MR [Socher *et al.*, 2013]. And each review has fine-grained labels (very positive, positive, neutral, negative, very negative), moreover, phrase-level annotations on all inner nodes are provided.

- SST-2: The same dataset as SST-1 but used in binary mode without neutral sentences.

- Subj: Subjectivity dataset is from [Pang and Lee, 2004], where the task is to classify a sentence as being subjective or objective.

- TREC: TREC is a dataset for question type classification task [Li and Roth, 2002]. The sentences are questions from 6 classes (person, location, numeric information, etc.).

### 4.2 Implementation Details

In the experiments, we use publicly available 300-dimensional Glove vectors that were trained on 42 billion words, moreover, the words are case-insensitive. For those words not present in the set of the pre-trained words, we just abandon them.

For model details, the number of hidden units of top Bi-LSTM (the last Bi-LSTM layer in dense Bi-LSTM module)

Table 1: Classification accuracy of DC-Bi-LSTM against other state-of-the-art models. The best result of each dataset is highlighted in **bold**. There are mainly five blocks: i) traditional machine learning methods; ii) Recursive Neural Networks models; iii) Recurrent Neural Networks models; iv) Convolutional Neural Net-works models; v) a collection of other models. **SVM**: Support Vector Machines with unigram features [Socher *et al.*, 2013] **NB**: Na-ive Bayes with unigram features [Socher *et al.*, 2013] **Standard-RNN**: Standard Recursive Neural Network [Socher *et al.*, 2013] **RNTN**: Recursive Neural Tensor Network [Socher *et al.*, 2013] **DRNN**: Deep Recursive Neural Network [Irsoy and Cardie, 2014] **LSTM**: Standard Long Short-Term Memory Network [Tai *et al.*, 2015] **Bi-LSTM**: Bidirectional LSTM [Tai *et al.*, 2015] **Tree-LSTM**: Tree-Structured LSTM [Tai *et al.*, 2015] **LR-Bi-LSTM**: Bidirectional LSTM with linguistically regularization [Qian *et al.*, 2016] **CNN-MC**: Convolutional Neural Network with two channels [Kim, 2014] **DCNN**: Dynamic Convolutional Neural Network with k-max pooling [Kalchbrenner *et al.*, 2014] **MVCNN**: Multi-channel Variable-Size Convolution Neural Network [Yin and Schütze, 2016] **DSCNN**: Dependency Sensitive Convolutional Neural Networks that use CNN to obtain the sentence representation based on the context representations from LSTM [Zhang *et al.*, 2016a] **BLSTM-2DCNN**: Bidirectional LSTM with Two-dimensional Max Pooling [Zhou *et al.*, 2016].

| Model | MR | SST-1 | SST-2 | Subj | TREC |
|---|---|---|---|---|---|
| SVM [Socher *et al.*, 2013] | - | 40.7 | 79.4 | - | - |
| NB [Socher *et al.*, 2013] | - | 41.0 | 81.8 | - | - |
| Standard-RNN [Socher *et al.*, 2013] | - | 43.2 | 82.4 | - | - |
| RNTN [Socher *et al.*, 2013] | - | 45.7 | 85.4 | - | - |
| DRNN [Irsoy and Cardie, 2014] | - | 49.8 | 86.6 | - | - |
| LSTM [Tai *et al.*, 2015] | - | 46.4 | 84.9 | - | - |
| Bi-LSTM [Tai *et al.*, 2015] | 81.8 | 49.1 | 87.5 | 93.0 | 93.6 |
| Tree-LSTM [Tai *et al.*, 2015] | - | 51.0 | 88.0 | - | - |
| LR-Bi-LSTM [Qian *et al.*, 2016] | 82.1 | 50.6 | - | - | - |
| CNN-MC [Kim, 2014] | 81.1 | 47.4 | 88.1 | 93.2 | 92.2 |
| DCNN [Kalchbrenner *et al.*, 2014] | - | 48.5 | 86.8 | - | 93.0 |
| MVCNN [Yin and Schütze, 2016] | - | 49.6 | 89.4 | 93.9 | - |
| DSCNN [Zhang *et al.*, 2016a] | 81.5 | 49.7 | 89.1 | 93.2 | 95.4 |
| BLSTM-2DCNN [Zhou *et al.*, 2016] | 82.3 | **52.4** | 89.5 | 94.0 | **96.1** |
| DC-Bi-LSTM (**ours**) | **82.8** | 51.9 | **89.7** | **94.5** | 95.6 |

is 100, for the rest layers of dense Bi-LSTM module, the number of hidden units and layers are 13 and 15 respectively.

For training details, we use the stochastic gradient descent (SGD) algorithm and Adam update rule with shuffled mini-batch. Batch size and learning rate are set to 200 and 0.005, respectively. As for regularization, dropout is applied for word embeddings and the output of average pooling, besides, we perform L2 constraints over the soft-max parameters.

### 4.3 Results

Results of DC-Bi-LSTM and other state-of-the-art models on five benchmark datasets are listed in Table 1. Performance is measured in accuracy. We can see that DC-Bi-LSTM gets consistently better results over other methods, specifically, DC-Bi-LSTM achieves new state-of-the-art results on three datasets (MR, SST-2 and Subj) and slightly lower accuracy than BLSTM-2DCNN on TREC and SST-1. In addition, we have the following observations:

- Although DC-Bi-LSTM is a simple sequence model, but it defeats Recursive Neural Networks models and Tree-LSTM, which relies on parsers to build tree-structured neural models.

- DC-Bi-LSTM obtains significant improvement over the counterparts (Bi-LSTM) and variant (LR-Bi-LSTM) that uses linguistic resources.

- DC-Bi-LSTM defeats all CNN models in all datasets.

Above observations demonstrate that DC-Bi-LSTM is quite effective compared with other models.

### 4.4 Discussions

Moreover, we conducted some experiments to further explore DC-Bi-LSTM. For simplicity, we denote the number of hidden units of top Bi-LSTM (the last Bi-LSTM layer in dense Bi-LSTM module) as $th$, for the rest layers of dense Bi-LSTM module, the number of hidden units and layers are denoted as $dh$ and $dl$ respectively. We tried several variants of DC-Bi-LSTM with different $dh$, $dl$ and $th$, The results are shown below.

**(1) Better parameter efficiency**

Better parameter efficiency means obtaining better performance with equal or less parameters. In order to verify DC-Bi-LSTM has better parameter efficiency than Bi-LSTM, we limit the number of parameters of all models at 1.44 million (1.44M) and conduct experiments on SST-1 and SST-2. The results are shown in Table 3.

The first model in Table 3 is actually Bi-LSTM with 300 hidden units, which is used as the baseline model, and the results are consistent with the paper [Tai *et al.*, 2015]. Based on the results of Table 3, we get the following conclusions:

- DC-Bi-LSTM improves parameter efficiency. Pay attention to the second to the fifth model, compared with baseline model, the increase on SST-1(SST-2) are 0.4% (1.2%), 1.8% (1.3%), 2.7% (2.5%) and 1% (1.6%), respectively, with the parameters not increased, which demonstrates that DC-Bi-LSTM models have better parameter efficiency than base-

Table 3: Classification accuracy of DC-Bi-LSTM with different hyper parameters. we limit the parameters of all models at 1.44M in order to verify DC-Bi-LSTM models have better parameter efficiency than Bi-LSTM.

| $dl$ | $dh$ | $th$ | Params | SST-1 | SST-2 |
|---|---|---|---|---|---|
| 0 | 10 | 300 | 1.44M | 49.2 | 87.2 |
| 5 | 40 | 100 | 1.44M | 49.6 | 88.4 |
| 10 | 20 | 100 | 1.44M | 51.0 | 88.5 |
| 15 | 13 | 100 | 1.40M | **51.9** | **89.7** |
| 20 | 10 | 100 | 1.44M | 50.2 | 88.8 |

Table 4: Classification accuracy of DC-Bi-LSTM with different hyper parameters. we increase $dl$ gradually and fix $dh$ at 10 in order to verify that increasing $dl$ does improve performance of DC-Bi-LSTM models.

| $dl$ | $dh$ | $th$ | Params | SST-1 | SST-2 |
|---|---|---|---|---|---|
| 0 | 10 | 100 | 0.32M | 48.5 | 87.5 |
| 5 | 10 | 100 | 0.54M | 49.4 | 88.1 |
| 10 | 10 | 100 | 0.80M | 49.5 | 88.4 |
| 15 | 10 | 100 | 1.10M | **50.6** | **88.8** |
| 20 | 10 | 100 | 1.44M | 50.2 | **88.8** |

Table 5: Classification accuracy of DC-Bi-LSTM with different hyper parameters. we increase $dh$ gradually and fix $dl$ at 10 in order to explore the effect of $dh$ models.

| $dl$ | $dh$ | $th$ | Params | SST-1 | SST-2 |
|---|---|---|---|---|---|
| 10 | 0 | 100 | 0.32M | 48.5 | 87.5 |
| 10 | 5 | 100 | 0.54M | 49.2 | 88.3 |
| 10 | 10 | 100 | 0.80M | 49.5 | 88.4 |
| 10 | 15 | 100 | 1.10M | 50.2 | 88.4 |
| 10 | 20 | 100 | 1.44M | **51.0** | **88.5** |

line model

- DC-Bi-LSTM models are easy to train even when the they are very deep. We can see that DC-Bi-LSTM with depth of 20 (the fifth model in Table 3) can be successfully trained and gets better results than baseline model (50.2% vs. 49.2% in SST-1, 88.8% vs. 87.2% in SST-2). In contrast, we trained deep stacked LSTM on SST-1, when depth reached more than five, the performance (For example, 30% when the depth is 8, which drops 19.2% compared with baseline model) drastically decreased.

- The fifth model performs worse than the fourth model, which indicates that too many layers will bring side effects when limiting the number of parameters. One possible reason is that more layer lead to less hidden units (to ensure the same number of parameters), impairing the ability of each Bi-LSTM layer to capture contextual information.

**(2) Effects of increasing depth ($dl$)**

In order to verify that increasing $dl$ does improve performance of DC-Bi-LSTM models, we increase $dl$ gradually and fix $dh$ at 10. The results on SST-1 and SST-2 are shown in Table 4.

The first model in Table 4 is actually Bi-LSTM with 100 hidden units, which is used as the baseline model. Based on the results of Table 4, we can get the following conclusions:

- It is obvious that the performance of DC-Bi-LSTM is positively related to $dl$. Compared with baseline model, DC-Bi-LSTM with $dl$ equal to 5, 10, 15 and 20 get improvements on SST-1 (SST-2) by 0.9% (0.6%), 1.0% (0.9%), 2.1% (1.3%) and 1.7% (1.3%) respectively.

- Among all models, the model with $dl$ equal to 15 works best. As $dl$ continues to increase, the accuracy does not further improve, nevertheless, there is no significant decrease.

- Compared with the first model in Table 3, the fourth model here uses less parameters (1.10M vs. 1.44M) but performs much better (50.6% vs. 49.2% in SST-1, 88.8% vs. 87.2% in SST-2), which further proves that DC-Bi-LSTM models have better parameter efficiency.

**(3) Effects of adding hidden units ($dh$)**

In this part, we explore the effect of $dh$. The number of layers in dense Bi-LSTM module ($dl$) is fixed at 10 while the number of hidden units ($dh$) is gradually increased. The results on SST-1 and SST-2 are shown in Table 5.

Similarly, we use Bi-LSTM with 100 hidden units as baseline model (the first model in Table 5). Based on the results of Table 5, we can get the following conclusions:

- Comparing the first two models, we find that the second model outperforms baseline by 0.7% on SST-1 and 0.8% on SST-2, which shows that even if $dh$ is equal to 5, DC-Bi-LSTM are still effective.

- As $dh$ increases, the performance of DC-Bi-LSTM steadily increases. One possible reason is that the ability of each layer to capture contextual information is enhanced, which eventually leads to the improvement of classification accuracy.

## 5 Conclusion & Future Work

In this work, we propose a novel multi-layer RNN model called Densely Connected Bidirectional LSTM (DC-Bi-LSTM) for sentence classification tasks. DC-Bi-LSTM alleviates the problems of vanishing-gradient and overfitting and can be successfully trained when the networks are as deep as dozens of layers. We evaluate our proposed model on five benchmark datasets of sentence classification, experiments show that our model obtains significant improvements over the traditional Bi-LSTM and gets promising performance in comparison with the state-of-the-art approaches. As future work, we plan to apply DC-Bi-LSTM in the task of machine translation and dialog system to further improve their performance, for example, replace the Bi-LSTM encoder with dense Bi-LSTM module.